\documentclass[a4paper,fleqn]{cas-dc}

\usepackage{lineno,hyperref}
\usepackage{algorithm}
\usepackage{color,soul}
\usepackage{multirow}
\usepackage{amsmath}
\usepackage{tabularx}
\usepackage{lipsum}
\usepackage[noend]{algpseudocode}
\usepackage[utf8]{inputenc}
\usepackage[authoryear,longnamesfirst]{natbib}
\def\tsc#1{\csdef{#1}{\textsc{\lowercase{#1}}\xspace}}
\tsc{WGM}
\tsc{QE}
\tsc{EP}
\tsc{PMS}
\tsc{BEC}
\tsc{DE}

\begin{document}
\let\WriteBookmarks\relax
\def\floatpagepagefraction{1}
\def\textpagefraction{.001}

\shorttitle{A machine learning approach for predicting attrition patterns}
\shortauthors{Hamed Fayyaz et~al.}
\title [mode = title]{Who will Leave a Pediatric Weight Management Program and When? - A machine learning approach for predicting attrition patterns}

\author[1]{Hamed Fayyaz}[orcid=0000-0003-0053-5112]

\cormark[1]
\ead{fayyaz@udel.edu}
\author[2]{Thao-Ly T. Phan}
\author[2]{H. Timothy Bunnell}
\author[1]{Rahmatollah Beheshti}

\cortext[cor1]{Corresponding author}

\begin{abstract}
Childhood obesity is a major public health concern. Multidisciplinary pediatric weight management programs are considered standard treatment for children with obesity and severe obesity who are not able to be successfully managed in the primary care setting; however, high dropout rates (referred to as attrition) are a major hurdle in delivering successful interventions. Predicting attrition patterns can help providers  reduce the attrition rates. Previous work has mainly focused on finding static predictors of attrition using statistical analysis methods. In this study, we present a machine learning model to predict (a) the likelihood of attrition, and (b) the change in body-mass index (BMI) percentile of children, at different time points after joining a weight management program. We use a five-year dataset containing information about 4,550 children of diverse backgrounds receiving treatment at four pediatric weight management programs in geographically different regions of the country. Our models show strong prediction performance as determined by high AUROC scores across different tasks (average AUROC of 0.75 for predicting attrition, and 0.73 for predicting weight outcomes). Additionally, we report the top features that can predict attrition and weight outcomes at different times after joining the weight management program.

\end{abstract}



\maketitle

\section{Introduction}
Childhood obesity is a major public health concern across the world. The prevalence of overweight and obesity among children and adolescents aged 2-19 has risen dramatically from  29\% in 2000 to just over 32\% in 2012 in the United States \citep{skinner2014prevalence}. Childhood obesity increases the risk of diabetes, cardiovascular disease, and cancer, predominantly as a result of a substantially greater risk of adult obesity \citep{10.1001/jama.2014.732}, but also causes significant morbidity in childhood.  Multi-disciplinary clinical weight management  programs (WMPs) are recommended for children with obesity who fail to improve with management in the primary care setting but these programs often require a moderate to high intervention dose delivered over an extended period \citep{10.1111/ijpo.12733}. Children and their families who attend more
intervention sessions and remain enrolled in care for longer periods achieve the greatest improvements in
weight and health \citep{jamapediatrics.2017.2960}. However, due to a variety of reasons, such as dissatisfaction with the
intervention progress or logistical issues with attending the programs, many families (as much as 80\% \citep{10.1111/ijpo.12733}) discontinue attending WMPs prematurely (a phenomenon referred to as
``attrition''). Besides not treating the disease effectively, a failed weight loss attempt may also lead to frustration, discouragement, and learned helplessness \citep{Ponzo}. Attrition can also be challenging for healthcare systems needing to ensure effective delivery of their services and efficient
utilization of their often limited resources.   

Prior studies have explored reasons for and predictors of attrition \citep{10.1111/ijpo.12733, Jiandani, Moran, Ponzo, Altamura2018}. These studies have found several predictors of attrition from WMPs, such as sex, age, ethnicity, psychosocial factors \citep{phan2018impact}, and initial weight loss \citep{Batterham, Jiandani, Moroshko}. However, there is no consensus on how these various factors contribute to attrition, and what strategies can be employed to reduce attrition.  Another limitation of prior work relates to using only  static features (exposures) to study attrition, without explicitly considering temporal features, such as weight trajectories. Considering weight trajectories is pivotal in studying obesity patterns and also may be helpful in predicting attrition \citep{Kushner20}.

In this study, we present a machine learning model to address the current limitations and gaps in the field and apply our method to a large dataset, representing four pediatric WMP sites within the Nemours Children's Health System (a large pediatric health  system in the US).  Specifically, we present a deep (neural network) model with two separate components for analyzing the static and dynamic input features, extracted from the electronic health records (EHRs) of children attending the WMPs. To improve the overall predictive performance of our models,  we use a "multi-task learning" approach by combining the two prediction tasks (i.e., predicting attrition and weight outcomes) such that one model generates two values for these tasks. In our study, attrition prediction refers to predicting the time (number of weeks after the baseline visit) of the last visit, and the weight outcome prediction task refers to predicting the change in BMI percentile (BMI\%) of patients at their last visit to the WMP (classified as a decrease in BMI\% or not).  Besides multi-task learning, our model also follows a "transfer learning" design, by training our model on various lengths of observation  and  prediction windows \citep{10.1111/ijpo.12733}. Following the initial training of the model on these different settings, it is then fine-tuned (retrained) on a final target task to report the desired outcomes. Specifically, the main contributions of our study are:

\begin{itemize}
 \item We have collected one of the largest and most comprehensive datasets dedicated to study attrition.  The dataset represents 4,550 children from four WMP sites within a large pediatric healthcare system,  The dataset includes standard items from the EHR (e.g. medical diagnoses) and additional  lifestyle and psychosocial factors collected by the WMPs in the EHR as part of routine clinical care.  The dataset also includes temporal factors that may be important such as time between visits and BMI\% trajectories.
 \item We present a deep model for predicting when attrition occurs and patients' BMI\% change at that time. We use a multi-task and transfer learning approach for improving the performance of our model, facilitating its deployment in real settings without large training data.
 \item We compare our model against the commonly used logistic regression and a state of the art survival analysis method and show that it achieves better results in predicting attrition. 
 \item We study the role of various input features in predicting attrition and successful BMI\% change. This increases the interpretability of our model and makes our findings more actionable. 
\end{itemize}

\section{Related Work}

Studying attrition is similar to studying other healthcare problems such as visit attendance and treatment adherence in clinical settings.  Attendance prediction generally aims to predict whether a patient will show up for a scheduled appointment or not. Attendance prediction has been used to identify the important factors predicting visit attendance in various fields such as  rehabilitation \citep{SABIT2008819,HAYTON2013401}, psychiatric care \citep{mitchell_selmes_2007}, and primary care \citep{giunta2013factors, kheirkhah2015prevalence}. On the other hand, adherence prediction aims to predict whether a patient will be compliant with their treatment plan (e.g. taking prescribed medicines). Treatment and medication adherence have been studied for conditions such as tuberculosis \citep{killian2019learning}, heart failure \citep{son2010application}, and schizophrenia \citep{son2010application}. There are also other studies outside the healthcare domain that relate to our work. Among these related studies, one is "churn" prediction often used in economic domains to predict engagement patterns of individuals (such as customers and employees) to increase their retention. Churn prediction  has been also well researched in the fields of banking \citep{ali2014dynamic}, video games \citep{kawale2009churn, hadiji2014predicting}, and telecommunication \citep{huang2012customer}.

While the studies discussed so far closely relate to the field of attrition, a major distinguishing aspect of attrition studies is the chronicity of the conditions being treated and the commitment to an intervention or treatment required for success. Examples of conditions where attrition patterns have been studied include mental health conditions \citep{linardon2020attrition}, sleep disorders \citep{doi:10.1080/15402002.2010.487457}, and addiction disorders \citep{info:doi/10.2196/jmir.2336}. Attrition from WMPs, in particular, has been studied in different settings, mainly using traditional methods such as linear and logistic regression \citep{Jiandani, Altamura2018, Ponzo, perna2018path}. Linear and logistic regression assume that the relationship between the independent and the dependent variables are linear, which may not be correct. Using these standard methods, researchers have proposed various predictors of attrition, such as psychological \citep{Altamura2018},  \citep{Jiandani}, sociodemographic, and anthropometric \citep{Ponzo} factors, and initial weight-loss \citep{perna2018path}. Among this family of studies, Batterham et al. \citep{Batterham} were the only group who used shallow decision trees to predict attrition in dietary weight loss trials using demographic and early weight change characteristics. Their approach, however, did not use weight trajectories or other temporal patterns.

\section{Dataset}

Our dataset includes all children 0-21 years of age who visited one of the four WMPs of the Nemours Children's Health system between 2007 and 2021. The four WMPs serve patients in the US states of Delaware, Florida, Maryland, New Jersey, and Pennsylvania. For each patient, data from an internal dataset collected by the providers inside the WMPs were linked to EHR data, capturing their health records when interacting with the entire healthcare system. The internal dataset collected at the WMPs was specifically designed to capture important covariates generally missing in EHRs, such as psychosocial determinants of health and lifestyle behaviours (see Table \ref{tab:cohort}). The EHR dataset was the Nemours portion of the large PEDSnet data repository and included rigorously validated EHR variables including medical conditions, anthropometrics, visits, and demographics \citep{Forrest}. PEDSnet is a multi-speciality network that conducts observational research and clinical trials across multiple children's hospital health systems. The final dataset was anonymized, and our study was approved by the Nemours Institutional Review Board.

\begin{table*}[htbp]
 \caption{Characteristics of the study cohort. }
 \label{tab:cohort}
\begin{tabular}{p{0.25\textwidth}p{0.7\textwidth}}
\hline
Variable & Description\\\hline
Sex                        & Male(2,122),          Female(2,428)\\
Age &  Range=(1–19), Mean=10.5 \\
Race      & Asian(57), White(1,767), Black(1,219), Other(1,462) \\      
Ethnicity & Hispanic(1,680), Non-Hispanic(2,844)      \\ 
Time btw  visits  & Mean=7 weeks \\
BMI\% (per visit) & Mean BMI-for-age percentile at baseline visit=98 \\
Insurance type & Medicaid(1,998),   Private(1,570)\\
Food insecurity\textsuperscript{*}  & Often or  Sometimes true Mean item-1(646), item-2(427)  \\
Lifestyle score\textsuperscript{\dag}  & Range=(3–47), Mean=38.95\\
PSC-17\textsuperscript{\ddag} & Range=(3–33), Mean=9\\
WMP visit type & Nutrition(4,293),
 medical(12,927), psychology(1,734), exercise(3,028)\\

Diagnosis codes &  24 most commonly reported conditions  (e.g., diabetes) \\\hline
\end{tabular}
\footnotesize
\textsuperscript{*} as measured by the validated 2-item Hunger Vital Sign\citep{huger2009}. \textsuperscript{\dag} based on 12 evidence-based items about diet, activity, sleep, and hunger  (each scored on a 4-point Likert scale with  a total score of 12-48). \textsuperscript{\ddag} Pediatric Symptom Checklist \citep{y1994pediatric}, a validated 17-item screening tool (total score 0-34).
\end{table*}

\begin{figure*}[]
 \centering
 \includegraphics[width=1\linewidth]{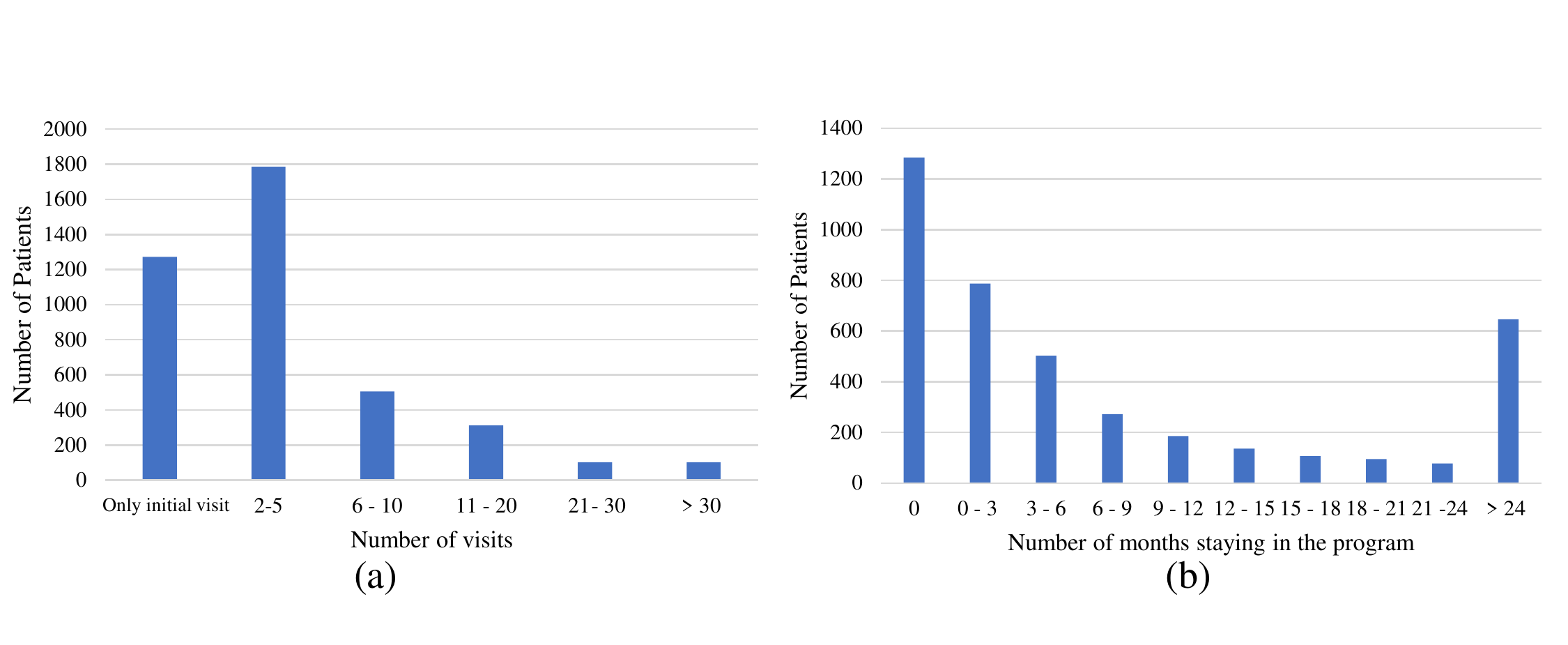}
 \caption{The distribution of (a) the number of visits and (b) the number of months staying in the WMP across the patients in our cohort.}
 \label{fig:visitcountandlength}
\end{figure*}

The final cohort included 4,550 children (with 26,895 total WMP visits) of diverse backgrounds (27\% Black, 37\% Hispanic, 48\% with Medicaid) with a mean BMI\% (BMI percentile) of 98. For the BMIs, we use age- and sex-adjusted BMI\% above the 95th percentile as defined by the Centers for Disease Control and Prevention (CDC) \citep{centersprevention_2017}. BMI\% is recorded and fed to the model whenever available (for the baseline visit and the majority of the follow-up visits) and the sequence of BMI\%s from baseline to the end time point of the particular prediction window is included as a dynamic variable.  For each child, we included 18 features capturing demographic, psychosocial, lifestyle, anthropometric, medical comorbidity, and visit  variables. Table \ref{tab:cohort} shows a list of these variables and their values. Our variables include two risk scores, capturing evidence-based lifestyle (diet, sleep, physical activity, sedentary activity, and hunger) and psychosocial  (mental health and behavioural) factors.  In our dataset, 28\% of the children had only one WMP visit, and  15\% had more than ten visits. Figure \ref{fig:visitcountandlength} shows the overall distributions of the number of visits and the duration of WMP attendance in months.

To bucketize the longitudinal EHR data, we combined visits over 15-days time periods.  We examined medical diagnoses codes and medication tables, from the available EHR data. Any condition observed at least once during the time window was denoted by 1 in the new sequence, and the measurements were averaged over the time window. If there were no visits for a patient in a time window, the corresponding vector for that period was set to all zeros. We also excluded rare diagnosis codes (i.e., the codes that appeared in less than $2\%$ of patients), which reduced the total number of diagnosis codes from 435 to 24. We used one-hot encoding for the categorical values and normalized the continuous values by performing a min-max scaling on all features.  

\section{Method}
As discussed earlier, in this study we considered two related predictive tasks, i.e., attrition prediction and weight outcome prediction. We considered these two prediction tasks in a binary classification framework. In the attrition prediction task, patients who dropped out before the prediction window were considered positive cases. In the outcome prediction task, any decrease in the child's BMI\% was considered a success. Accordingly, we defined the patients whose BMI\% in the prediction window was lower than their BMI\% at the time of their baseline visit to the WMP as negative cases. The rest of the patients (i.e., those whose BMI\% remained the same or increased during the observation window) were considered as positive cases. 

We used flexible observation and prediction windows for both tasks, where the start of the observation window was always fixed at the first WMP visit with a rolling end considered for the end of the window. The start of the prediction window was considered as the end of the observation window with a flexible end.  We note that  this type of flexibility in defining observation and prediction windows makes our models more practical in clinical settings, where a provider needs to know which child will leave at what time following the start of the program.



\begin{figure}[h]
 \centering
 \includegraphics[width=0.95\linewidth]{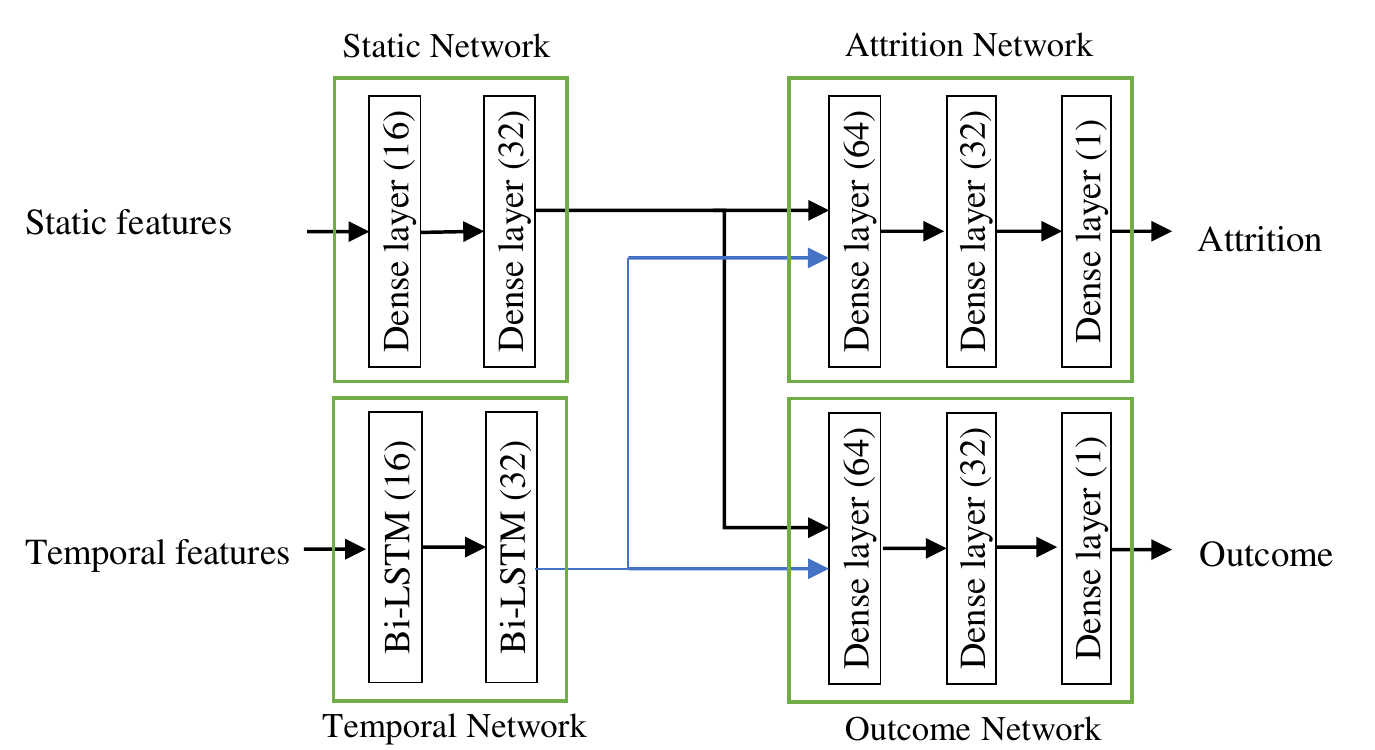}
 \caption{The proposed architecture for our model. The model receives static and temporal features and predicts the time of last visit (attrition) and the BMI\% change outcome at that time (Weight outcome). }
 \label{fig:model}
\end{figure}

For implementing the two prediction tasks, we propose a deep neural network architecture, specifically designed to address our problem's needs. In order to fully utilize the static (e.g., demographics) and temporal features (e.g., measurements) in our data, our architecture is designed with the following four components as shown in Figure \ref{fig:model}. Two initial components are shared between the two tasks of attrition and outcome predictions and the other two are task-specific components. The first component is a two-layer fully-connected network for processing the static features.  The second is a two-layer bi-directional long short-term memory (Bi-LSTM) network for processing the temporal features. Bi-LSTM structures are similar to the common LSTMs, but they consist of two LSTMs units: one taking the input in a forward direction (e.g., left to right), and the other in a backwards direction (e.g., right to left), thus improving the  available context for the model \citep{schuster1997bidirectional}.  The third component is a  three-layer fully connected network for combining the extracted feature vectors from the first two parts and predicting the attrition time.  Finally, the fourth component is a three-layer fully-connected network for combining the extracted feature vectors from the first two parts and predicting the outcome. We used dropout and batch normalization layers after all of the layers mentioned above. For training the model, we used binary cross-entropy as the loss function, defined as:

\begin{equation*}
\frac{1}{N} \sum_{i=1}^{N}- ( y_{i}*log(p_{i}) + (1-y_{i})*log(1-p_{i})),
\end{equation*}
\noindent
where, N is the total number of samples, $y_{i}$ is the ground truth for the $i$th sample, $p_i$ is the probability of belonging to the positive class , and $(1-p_{i})$ is the probability of belonging to the negative class. In the attrition task, we train a model to discriminate between the patients who dropped out and patients who stayed in the WMP. In the outcome task, we train a model to discriminate between patients who successfully decreased their BMI\% and those who had no change or a BMI\% increase.

\begin{figure*}[h]
 \centering
 \includegraphics[width=0.85\linewidth]{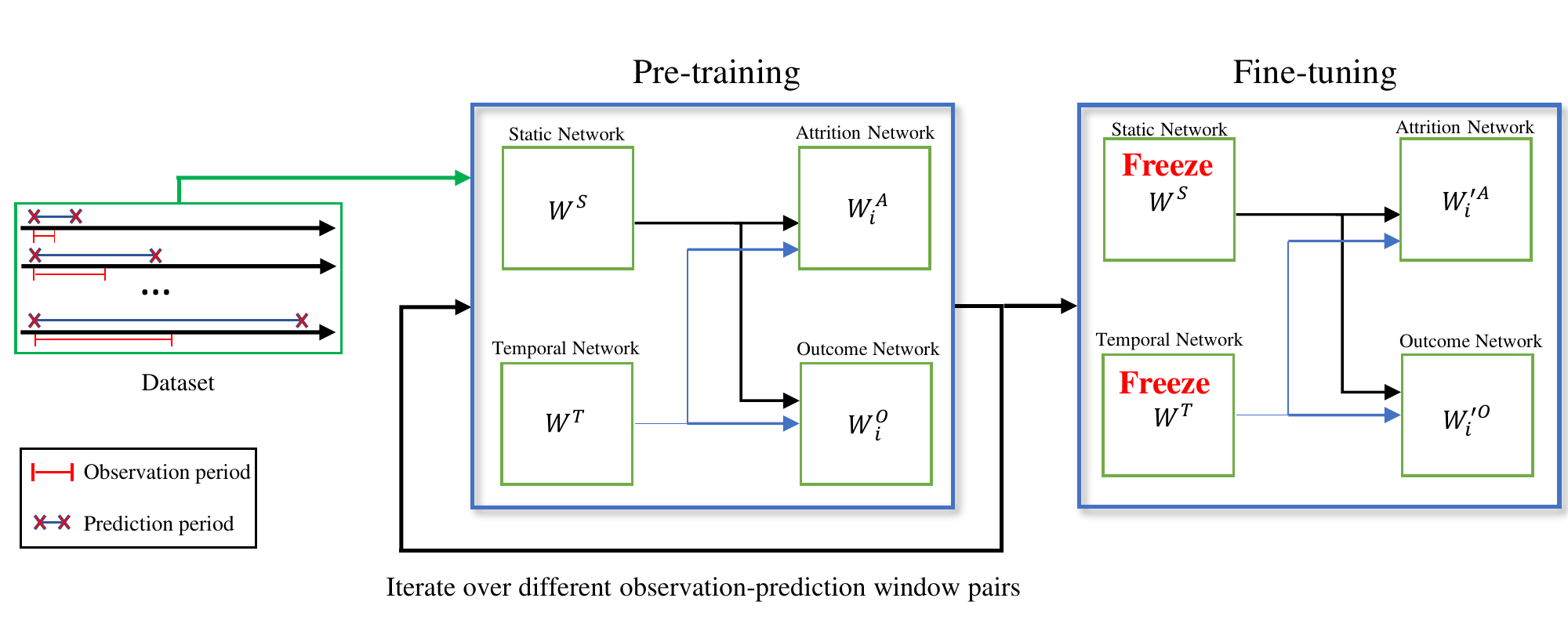}
 \caption{Proposed training process  using the multi-task and transfer learning themes.  $W^S$, $W^A$, $W^T$, and $W^O$ are the weights of the static, attrition, temporal, and outcome sub-networks, respectively. We extract the features and the corresponding labels for each patient based on the rolling observation and prediction windows and feed them to the network. After the pretraining of a general model, we then initialize the weights of the specialized models with the weights from the general model. For each specific observation and prediction window setting, the model is then fine-tuned using only the relevant samples. $W_j$ shows the $j$th fine-tuned configuration.}
 \label{fig:transferlearning}
\end{figure*}

To improve the overall performance of our models and accommodate the relatively small size of our training data, we use a "multi-task learning" approach for designing our model. In our design, we use a hard parameter sharing approach, which is a common multi-task learning  method \citep{Caruana93multitasklearning}, by sharing the first two components between the two tasks. This way, sharing the parameters of the two initial parts of the model can improve the performance for both predictions tasks. Additionally, following a ``transfer learning'' theme, we iteratively  train our model on all sliding observation and prediction windows, and then fine-tune the model for each specific window settings. Figure \ref{fig:transferlearning} shows the training process and the way that the four components are involved in our multi-task and transfer learning themes.   Additionally, Algorithm \ref{alg:train} presents a high-level pseudocode of our customized model training process. In this algorithm,  each model is trained using the data  from a specific observation and prediction window, initialized by the weights from the previous window settings. The procedure in this algorithm receives the input data ($X$), labels for the attrition and outcome tasks ($Y_A$ and $Y_O$), and the list of observation and prediction windows. It returns a list of fine-tuned models. The model presented in this paper was implemented using  Keras \citep{chollet2015keras} inside the  TensorFlow \citep{abadi2016tensorflow} framework. Our code is publicly available on GitHub\footnote{\url{https://github.com/healthylaife/WM\_attrition}}.  

\begin{algorithm}
\caption{Training pseudocode}\label{alg:train}
\begin{algorithmic}[1]
\State \textbf{Input:} X, $Y_A$, $Y_O$, Observation-Prediction window list
\State \textbf{Output:} Trained models list 
\State Preprocess X
\State PM = random (PM) \Comment{Random initialization of the pretrained model}
\For{Every observation-prediction window pair}
\State $X^\prime$ = Select cohort based on the windows
\State M = PM \Comment{Load weights from the previous PM to model M}
\State Pretrain M with $X^\prime$ 
\State PM = M \Comment{Save M as the current pretrained model PM}
\State Freeze the first two components in M
\State Fine tune the rest of M
\State Add M to trained models list
\EndFor
\State \textbf{return} Trained models list
\end{algorithmic}
\end{algorithm}

\section{Experiments and Discussion}
To measure the prediction performance of our model, we report accuracy, precision, recall, area under the receiver operating characteristic (AUROC), area under the precision-recall curve (AUPRC), and baseline AUPRC. The last measure (baseline AUPRC) is the proportion of positive examples in our data.  The performance of the proposed model in the outcome and attrition prediction tasks for a series of observation  and prediction windows (observation window = 1, 2, 3, 4, 6, and 9 months; and prediction window = 1.5, 3, 4.5, 6, 9, 13.5 months) are shown in Table \ref{tab:result1} and Table \ref{tab:result2}. This specific set of observations and windows were selected based on prediction windows used in prior studies \citep{Moran, Jiandani} and based on the distribution of our data. The results show that the performance of the models has a direct relationship with the length of the observation window, i.e., as we expand the observation window through different tasks, the model's performance also increases.

\begin{table*}[h]
\centering
\caption{Results for the attrition prediction task using our proposed method. Observation and prediction windows are shown in months.}
\label{tab:result1}
\begin{tabular}{llllllll}
\hline
Observation & Predication & Precision & Recall & AUROC & AUPRC & B.AUPRC \\ \hline
1            & 1.5          & 0.48      & 0.74   & 0.62  & 0.49  & 0.40    \\
2            & 3            & 0.65      & 0.63   & 0.68  & 0.64  & 0.50    \\
3            & 4.5          & 0.76      & 0.69   & 0.75  & 0.76  & 0.58    \\
4            & 6            & 0.81      & 0.68   & 0.78  & 0.84  & 0.63    \\
6            & 9            & 0.87      & 0.70   & 0.80  & 0.89  & 0.69    \\
9            & 13.5        & 0.91      & 0.78   & 0.84  & 0.94  & 0.76   
\end{tabular}
\end{table*}

\begin{table*}[h]
\centering
\caption{Results for the outcome prediction task using our proposed method. Observation and prediction windows are shown in months.}
\label{tab:result2}
\begin{tabular}{llllllll}
\hline
Observation & Predication & Precision & Recall & AUROC & AUPRC & B.AUPRC \\ \hline
1            & 1.5          & 0.21      & 0.65   & 0.54  & 0.23  & 0.19    \\
2            & 3            & 0.52      & 0.62   & 0.77  & 0.64  & 0.27    \\
3            & 4.5          & 0.44      & 0.68   & 0.74  & 0.58  & 0.28    \\
4            & 6            & 0.59      & 0.78   & 0.84  & 0.71  & 0.31    \\
6            & 9            & 0.42      & 0.64   & 0.71  & 0.59  & 0.31    \\
9            & 13.5         & 0.54      & 0.60   & 0.75  & 0.66  & 0.33   
\end{tabular}
\end{table*}

We compare our model against two separate baselines. The first is a general logistic regression (LR) method, which is a frequently used method in the literature \citep{shipe2019developing}. Similar to the common practice in the field, we aggregate the temporal variables by calculating the average values during a corresponding period. We train an LR model with L2 penalty and BFGS solver using the scikit-learn library \citep{pedregosa2011scikit} in Python.

\begin{figure*}[h]
 \centering
 \includegraphics[width=1\linewidth]{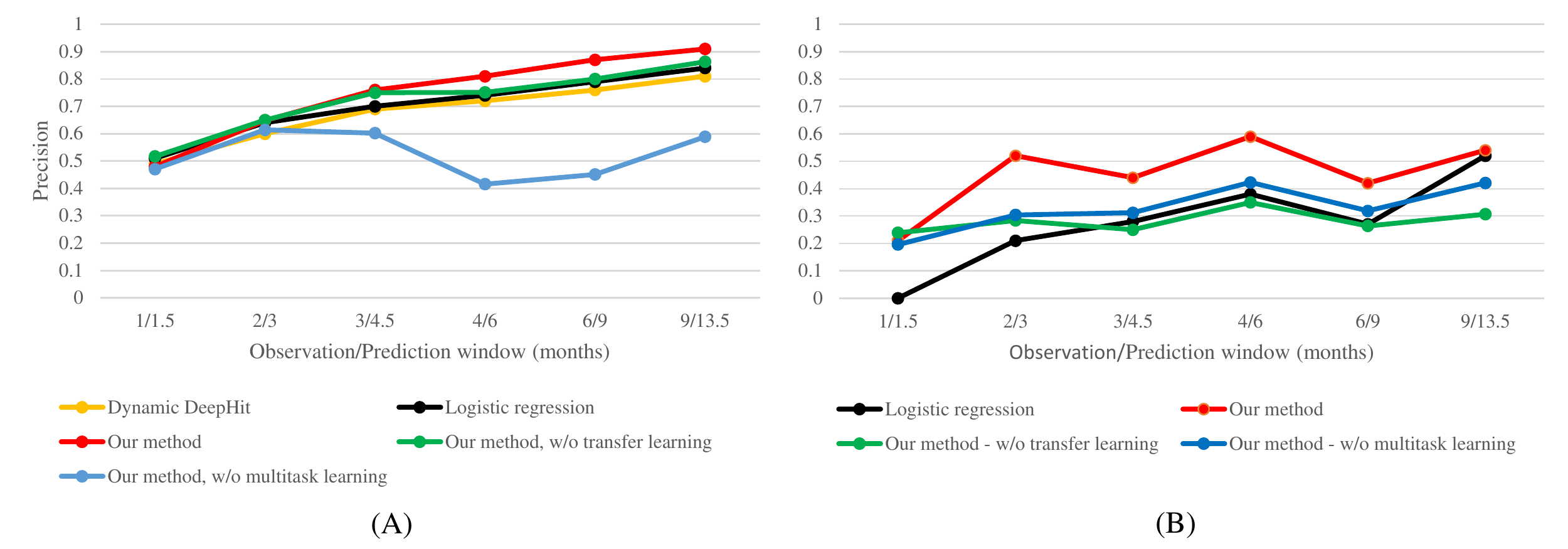}
 \caption{ Performance of our method versus the logistic regression and Dynamic DeepHit \citep{DynamicDeepHit} baselines as determined by precision in predicting (A) attrition, and (B) weight outcome (without the Dynamic DeepHit). }
 \label{fig:compare}
\end{figure*}

For the second baseline, we use another popular method for studying similar problems in biomedical domains, which takes a survival analysis approach. Survival analysis (time-to-event analysis) is widely used in many engineering, economics, and medicine \citep{clark2003survival} applications to  estimate the expected duration until a desired event occurs (e.g., drop out or death). We use a state of the art method for studying survival analysis, called  Dynamic DeepHit \citep{DynamicDeepHit}, which presents a deep neural network to learn the distribution of survival times. Dynamic DeepHit utilizes the available temporal data  to issue dynamically updated survival predictions. This survival analysis method only fits our attrition prediction task (and not the weight outcome prediction task). In an ablation analysis, we also study the role of the multi-task and transfer learning themes used in our design. To this end, we report the results of our method without (a) multi-task learning implementation (i.e., the two shared sub-networks and only one of the latter sub-networks for each task), and (b) transfer learning implementation (i.e., fine-tuning the models on the corresponding data to each observation-prediction window setting without pretraining). Figure \ref{fig:compare} shows how the results obtained from our method compares to the two baselines and the ablated version of our model.

Moreover, to study to what degree each input feature contributes to predicting the final outputs in each task, we used the popular Shapley additive explanations (SHAP) method implemented by the SHAP toolbox \citep{NIPS2017_7062, Scott}. Inspired by the Shapley values, this toolbox is a unified framework that assigns each input feature an importance value indicating its importance in the prediction task. A SHAP value indicates the degree to which a feature contributes to pushing the output from the base value (average model output) to the actual predicted value, so the higher value means the higher importance and contribution toward the outcome of the model. Here, we report the importance  values (as indicated by SHAP) for the top 5 features predicting each outcome and for each prediction/observation window.  These are shown in tables \ref{fig:SHAP1} and \ref{fig:SHAP2}.  Not surprisingly, the top features did vary some by the outcome and prediction/observation window, which demonstrates the importance of targeting different risk factors at different points in treatment.  However, age, the average time between the visits,  and patient's BMI\% trajectory were important features for determining both attrition and weight outcomes, which is consistent with previous studies \citep{Batterham, batterham2016predicting}.  Specifically, a patients' early weight loss is predictive of overall weight loss progress and the patients who have success with early weight loss seem to have a lower risk of attrition. Similar to our study, other studies \citep{Jiandani, batterham2016predicting} have also found that age is an important predictor of both attrition and weight outcomes, with children of younger ages having more success in WMPs. We have also found that the average time between WMP visits is a primary predictor for both tasks, with worse outcomes and higher attrition rates for patients who had a prolonged time between visits.  This is consistent with the literature and current guidelines recommending a certain number of contact hours and frequency of contact for success in WMPs. Finally, sociodemographics like race and ethnicity and sex, as well as food security and insurance status, are important predictors of both attrition and weight outcomes, consistent with previous studies demonstrating inequities in health outcomes between subgroups \citep{MARTIN200731}. Interestingly, other factors like medical diagnoses, medications, and lifestyle scores were less predictive of attrition and weight outcomes, which highlights the importance of identifying and supporting key groups based on sociodemographics, as well as ensuring frequent visits and early success with weight outcomes during treatment regardless of a patient's underlying conditions or lifestyle behaviors. We note that while we report the average SHAP values across all of the samples (children), our  model is most helpful when individual children are considered separately, and a provider can see the predictors that can be targeted for a particular child to prevent attrition or increase success with weight outcomes.  

\setlength{\tabcolsep}{0pt}
\begin{table*}[]
  \caption{The top five features predict attrition in various observation and prediction window settings, as determined by the SHAP values shown in parentheses.  BMI\% shows the BMI\% trajectory recorded during the observation window. Food ins:Food insecurity. Visits   int: Visits   intervals.}
\begin{tabularx}{\textwidth}{p{3cm}p{3cm}p{3cm}p{3cm}p{3cm}p{3cm}}
\hline
\multicolumn{6}{c}{Observation/Prediction window (in months)}                                                                                                      \\ \hline
1/1.5                  & 2/3              & 3/4.5                     & 4/6                       & 6/9                       & 9/13.5                    \\ \hline
Age (0.041)            & Insurance(0.015) & Visits   int.(0.023) & Visits   int.(0.020) & Age(0.063)                & Age(0.012)                \\
Ethnicity(0.006)    & BMI   \%(0.006)  & BMI   \%(0.020)           & Age(0.012)                & Insurance(0.009)          & Visits   int.(0.005) \\
Food ins.(0.005) & Race(0.005)      & Age(0.010)                & BMI   \%(0.012)           & Visits   int.(0.003) & BMI   \%(0.002)           \\
Sex(0.005)             & Age(0.005)       & Food   ins.(0.007)  & Insurance(0.007)          & Sex(0.002)                & Insurance(0.0002)         \\
Insurance(0.003)       & Sex(0.004)       & Insurance(0.005)          & Sex(0.006)                & BMI   \%(0.002)           & Lifestyle   score(0.0002) \\ \hline
\end{tabularx}
\label{fig:SHAP1}
\end{table*}

\begin{table*}[]
  \caption{The top five features predict the weight outcome in various observation and prediction window settings, as determined by the SHAP values shown in parentheses. BMI\% shows the BMI\% trajectory recorded during the observation window. Food ins:Food insecurity. Visits   int: Visits   intervals.}
\begin{tabularx}{\textwidth}{p{3cm}p{3cm}p{3cm}p{3cm}p{3cm}p{3cm}}
\hline
\multicolumn{6}{c}{Observation/Prediction window (in months)}                                                                                                               \\ \hline
1/1.5                  & 2/3                       & 3/4.5                     & 4/6                       & 6/9                       & 9/13.5                    \\ \hline
Age (0.010)            & Age(0.058)                & Insurance(0.020)          & Age(0.043)                & Age(0.021)                & Age(0.014)                \\
Race(0.005)            & Visits   int.(0.014) & Age(0.018)                & BMI   \%(0.029)           & Visits   int.(0.008) & BMI   \%(0.006)           \\
Insurance(0.004)       & Insurance(0.014)          & Ethnicity(0.012)     & Visits   int.(0.019) & Food   ins.(0.007)  & Visits   int.(0.003) \\
Lifestyle Score(0.003) & Sex(0.008)                & Race(0.011)               & Race(0.006)               & BMI   \%(0.005)           & Race(0.0004)              \\
Food ins.(0.003) & BMI   \%(0.006)           & visits   int.(0.010) & Sex(0.005)                & Sex(0.005)                & Sex(0.0003)               \\ \hline
\end{tabularx}
 \label{fig:SHAP2}
\end{table*}

The current study is limited in several ways. First, our dataset only includes the patients from one healthcare system. Still, our dataset is larger than similar ones used to study attrition and spans four geographically distinct sites in the Mid-Atlantic (Delaware, Pennsylvania, Maryland, New Jersey) and Southern (Florida) regions of the US.  
Additionally, our approach relies on discretizing future time into two-week windows. Considering attrition prediction as a regression task was an alternative natural choice. In our experiments, we have noticed that formulating our problem as a classification task yields better results. Additionally, as most follow-up visits are not scheduled in shorter than two-weeks intervals, one can still use our approach for continuous (any time in future) predictions. While our study focuses primarily on attrition from pediatric WMPs, our method should be applicable to adult obesity WMPs and other comparable problems such as mental health and addiction recovery programs.  As part of our future work, we plan to expand our model by including additional information from children's historical (before joining a WMP) records and also by including new data from other pediatric health systems. On the technical side, we plan to use a transformer-based architecture \citep{vaswani2017attention}, which offers state of the art sequence-to-sequence predictive modelling to check whether it improves our model's performance  further. Moreover, we aim to explicitly identify the temporal patterns (such as distinct shapes of the body-weight trajectory) that can predict (or stratify) attrition or outcome patterns.

\section{Conclusion}
In this study, we presented a deep neural network that includes Bi-LSTM elements for predicting attrition patterns and weight outcomes in pediatric weight management programs. We trained our model with a large electronic health records dataset collected from the Nemours Children's Health. Our model showed strong prediction performance as determined by AUROC scores across different tasks and superiority against other baseline methods, including logistic regression and a survival analysis method. Finally, we explored the importance of individual variables in predicting attrition and weight outcomes and found that sociodemographic variables as well as early weight trajectory and visit timing were important across both outcomes.

\bibliographystyle{cas-model2-names}
\bibliography{ref}
\end{document}